\pdfoutput=1
\documentclass[letterpaper, 10 pt, conference]{ieeeconf}  

\IEEEoverridecommandlockouts                              

\usepackage{graphics} 
\usepackage{graphicx}  
\newcommand\norm[1]{\left\lVert#1\right\rVert}
\usepackage[ruled]{algorithm2e}
\usepackage{amsmath,amssymb} 
\usepackage[caption=false]{subfig}
\usepackage{comment}
\usepackage{color}

\title{\LARGE \bf
Learning Monocular Visual Odometry \\ through Geometry-Aware Curriculum Learning
}

\author{Muhamad Risqi U. Saputra$^{1}$, Pedro P. B. de Gusmao$^{1}$, Sen Wang$^{2}$, Andrew Markham$^{1}$, and Niki Trigoni$^{1}$
\thanks{$^{1}$Muhamad Risqi U. Saputra, Pedro P. B. de Gusmao, Andrew Markham, and Niki Trigoni are with Department of Computer Science, University of Oxford, UK {\tt\small firstname.lastname@cs.ox.ac.uk}}%
\thanks{$^{2}$Sen Wang is with School of Engineering and Physical Sciences, Heriot Watt University,
        UK {\tt\small s.wang@hw.ac.uk}}%
}

\begin{document}
\bstctlcite{IEEEexample:BSTcontrol}

\maketitle
\thispagestyle{empty}
\pagestyle{empty}

\begin{abstract}

Inspired by the cognitive process of humans and animals, Curriculum Learning (CL) trains a model by gradually increasing the difficulty of the training data. In this paper, we study whether CL can be applied to complex geometry problems like estimating monocular Visual Odometry (VO). Unlike existing CL approaches, we present a novel CL strategy for learning the geometry of monocular VO by gradually making the learning objective more difficult during training. To this end, we propose a novel geometry-aware objective function by jointly optimizing relative and composite transformations over small windows via bounded pose regression loss. A cascade optical flow network followed by recurrent network with a differentiable windowed composition layer, termed CL-VO, is devised to learn the proposed objective. Evaluation on three real-world datasets shows superior performance of CL-VO over state-of-the-art feature-based and learning-based VO.

\end{abstract}

\section{INTRODUCTION}

Visual Odometry (VO) is the task of estimating an agent's pose and trajectory from a sequence of images. This problem has interested researchers from both robotics and computer vision communities for several decades. Conventional VO methods rely on finding feature correspondences between consecutive frames and leverage multi-view geometry technique \cite{Hartley2004}. Despite its good performances, these feature-based approaches are very sensitive to noise, outliers, and dynamic objects \cite{Saputra2018}. Typical approach to tackle these drawbacks is accomplished by manually fine-tuning the algorithm parameters for different cases. However, a new paradigm based on Deep Neural Networks (DNNs) has recently emerged to alleviate the manual tuning problems by directly learning the model parameters from the data. Results from \cite{Muller2017,Wang2017,Wang2017a,Costante2018,Tang2018Geo} show that deep learning-based VO can yield comparable accuracy to the conventional approaches.

State-of-the-art deep learning-based VO typically minimizes the relative transformation loss as the objective function \cite{Wang2017}. Minimizing frame-to-frame relative loss generally can provide reasonable trajectory estimation, but it does not guarantee the consistency of the composed transformation when integrating those relative estimates. Adding the compositional transformation loss in the objective function is a natural way to introduce the consistency to the network. However, our experiments suggest that training deep learning-based VO using compositional transformation loss is hard to converge. Our intuition is that it is too difficult for the network to learn directly the complex geometry of composing the 6 Degree-of-Freedom (DoF) camera poses since the error of the predictions can be largely accumulated. An intuitive way to alleviate the difficulty of training complex geometry problem is by starting the learning process from an easier geometry task and then gradually increasing the difficulty of the task.

The idea of learning from small or easy tasks and progressively increasing the difficulties has been studied in the context of Curriculum Learning (CL). Inspired by the cognitive process of humans and animals, Bengio et al. \cite{bengio2009} proposed CL as a strategy to improve the convergence speed and generalization ability of a machine learning model by learning through highly organized or meaningful order of examples. In this paper, we study whether a similar learning strategy can be applied for estimating the complex geometry of monocular VO. In particular, we propose a deep neural network framework with geometry-aware objective function for learning monocular VO in an end-to-end manner and employ the CL strategy to gradually learn the proposed objective from a simpler objective. Our specific contributions are listed as follows:
\begin{itemize}
\item We present the first curriculum learning strategy for learning the geometry problem of monocular visual odometry, by gradually making the learning objective more difficult during training.
\item We propose a novel geometry-aware objective function by jointly optimizing relative transformation and its composition over small windows via bounded pose regression loss.
\item We design a network architecture, dubbed CL-VO, which consists of cascade optical flow network and recurrent networks with a differentiable windowed composition layer.
\item We evaluate the proposed approach on three datasets (2 public and 1 self-collected) and show that our method significantly outperforms state-of-the-art feature-based and learning-based VO approaches.
\end{itemize}

\begin{figure*}[!ht]
    \centering
    \includegraphics[width=1.8\columnwidth]{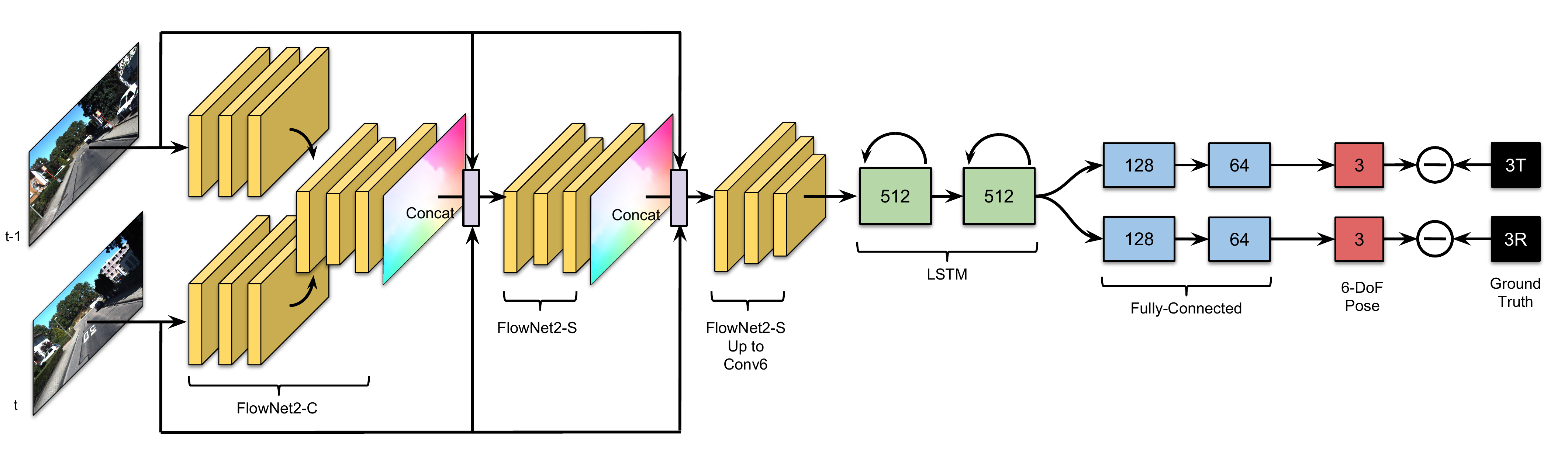}
    \caption{CL-VO architecture consists of cascade optical-flow networks followed by recurrent networks and fully connected layers.}
\label{fig:clvo_arch_rel}
\end{figure*}

\section{Related Work}

\subsection{Feature-based VO}

The feature-based VO pipeline generally starts by finding salient features, such as corners or blobs, and matching these features across frames. Using these feature correspondences, the camera ego motion can be estimated through the multiple view geometry principle. The first work on VO was proposed in 2004 by Nister in his landmark paper \cite{Nister2004f}. Subsequently, many variants were developed such as \cite{maimone2007,badino2013}, or VISO2 \cite{Geiger2011}. Estimating VO through feature-based approaches can be very accurate as it naturally follows the geometry of the camera and the captured scene. However, it can lack robustness due to noisy feature correspondences.

\subsection{Learning-based VO}

Since the advancement of DNNs, learning-based approaches are gaining more traction in solving computer vision tasks including VO. These approaches infer the camera pose by learning directly from real image data. Early works include Conda et al. \cite{Konda2015} which fed stereo images to a Convolutional Neural Network (CNN) to estimate the velocities and orientations of the camera through softmax-based classification. Flowdometry \cite{Muller2017} casted the VO problem as a regression problem by using FlowNet \cite{Dosovitskiy2016} to extract optical flow features and a fully connected layer to predict camera translation and rotation. The state-of-the-art approaches like DeepVO \cite{Wang2017} and ESP-VO \cite{Wang2017a} do not only utilize CNNs as the main building blocks, but also incorporate Recurrent Neural Networks (RNNs), to implicitly model the sequential motion dynamics of the image sequences. By not relying on finding feature correspondences, these approaches can yield more robust results in a variety of VO dataset \cite{Wang2017,Wang2017a,Tang2018Geo}.

\subsection{Curriculum Learning}

Curriculum Learning (CL) was proposed by Bengio et al. \cite{bengio2009} to formalize the idea of learning through a meaningful order of examples or concepts, which mimics how humans and animals learn. However, the basic idea of starting small or simple actually dates back to 1993 when Elman \cite{elman1993} successfully trained a DNN to recognize a simple grammar by increasing the complexity of the task. Bengio's work \cite{bengio2009} confirmed Elman's findings and showed that a well chosen CL strategy can improve the generalization ability of a DNN model. This idea was further improved by \cite{kumar2010} through Self-Paced Learning (SPL), in which the curriculum is learned during training rather than determined by prior knowledge. Jiang et al. \cite{jiang2015} then combined both idea of CL and SPL through Self-Paced Curriculum Learning (SPCL). SPCL takes into account both prior knowledge and the learning progress during training in constructing the curriculum. The application of CL and its improvement includes action detection \cite{jiang2014}, dictionary learning \cite{tang2012b}, domain adaptation \cite{tang2012a}, and object tracking \cite{supancic2013}, but none of them tackle VO estimation where it is more difficult to differentiate between easy and hard examples or tasks.

\section{Proposed Approach}

\subsection{Learning Ego-motion with DNNs}
The general approach to VO estimates a sequence of relative pose transformations $\{ \mathbf{\hat{p}}_{t-1}^{t}\} \subset \mathbf{SE}(3) $, from pairs of consecutive images $\{I_{t-1}$, $I_{t}\}$. The cumulative composition of these estimations generates a global trajectory with respect to the starting position i.e.
\begin{align}
\label{eq:trajectory_composition}
\textbf{\^{p}}^{t} & = \textbf{\^{p}}^{t}_{t-1} \oplus ... \oplus \textbf{\^{p}}^{2}_{1} \oplus \textbf{\^{p}}^{1} \;
\end{align}
where $\oplus$ represents the pose composition operation.

While conventional methods require the use of hand-crafted features and multiple view geometry techniques, DNN approaches work directly with raw image sequences by training the network in an end-to-end manner. Formally, given two concatenated images $\textbf{I}_{t-1,t} \in {\rm I\!R}^{2 \times ( w \times h \times c)}$ at times $t-1$ and $t$, where $w$, $h$, and $c$ are the image width, height, and channels respectively, DNNs learn the following mapping function to regress the 6-DoF camera pose:
\begin{align}
\label{eq:dnn_function}
\text{DNNs :} \{ ( {\rm I\!R}^{2 \times (w \times h \times c)} )_{1:N} \} \rightarrow \{ ( {\rm I\!R}^{6} )_{1:N} \} \;
\end{align}
where $N$ is the total number of consecutive image pairs.

\subsection{Enforcing Geometric Constraints}

During the training process, standard DNNs for VO estimation typically minimize relative transformation error between two consecutive frames. However, the ground truth pose is usually available as the composition of these relative transformations defining a sequence of global poses. In order to fully exploit both relative and composite transformation information, we need to jointly optimize these terms. Instead of directly placing relative and composite terms together, we propose to utilize the composed transformation as a constraint for the relative loss term. We only add the composite loss when its value at time $t$ is larger than it was at time $t-1$. This means that the network does not have to minimize the composite loss when the integration of relative poses at time $t$ yields more accurate absolute pose. Moreover, in order to reduce the accumulative errors, we only minimize the composite loss over small, bounded windows. We refer to this loss function as \textit{bounded pose regression loss}.

Equations (\ref{eq:bounded_loss})-(\ref{eq:regular_loss}) show this bounded loss, where $N$ is the number of images. $L_{rel}$ is the relative loss that measures pose errors between consecutive frames, while $L_{com}$ is the composite loss which accounts for errors over a small window. The coefficients $\alpha$ is used to balance both terms. 

The pose error defined in Equation \ref{eq:regular_loss} compares the estimated translation $\textbf{\^{t}}$ and rotation $\textbf{\^{r}}$ vectors (encapsulated in $\textbf{\^{p}}$) with their respective ground truth values.   
We also use $\delta$ and $\zeta$ to weigh the translation and rotation terms in relative loss as seen in \cite{Kendall2015,Wang2017}. 
\begin{align}
  L _{total} & = \sum_{t=1}^{N} \alpha L_{rel}  + (1-\alpha) L_{com}  \label{eq:bounded_loss} \\
  L_{rel} & = L\left( \textbf{\^{p}}^{t}_{t-1}\right) \label{eq:relative_loss} \\
  L_{com} & = 
  \begin{cases}
    L\left( \textbf{\^{p}}^{t}_{t-w} \right), &  \text{if } L\left( \textbf{\^{p}}^{t}_{t-w}\right) > L( \textbf{\^{p}}^{t-1}_{t-w-1}) \label{eq:conditioned_global_loss} \\
    0,              & \text{otherwise}
  \end{cases} \\
  L\left(\textbf{\^{p}}^{j}_{i}\right)  & = \delta \norm{\textbf{\^{t}}_{i}^{j} - \textbf{t}_{i}^{j}}^2 + \zeta \norm{\textbf{\^{r}}_{i}^{j} - \textbf{r}_{i}^{j}}^2 \label{eq:regular_loss}
\end{align}

\subsection{Geometry Aware Curriculum Learning}
\label{sec:gacl}

\begin{figure}[!ht]
    \centering
    \includegraphics[width=4.1cm,trim=1cm .0cm 1cm .cm,clip]{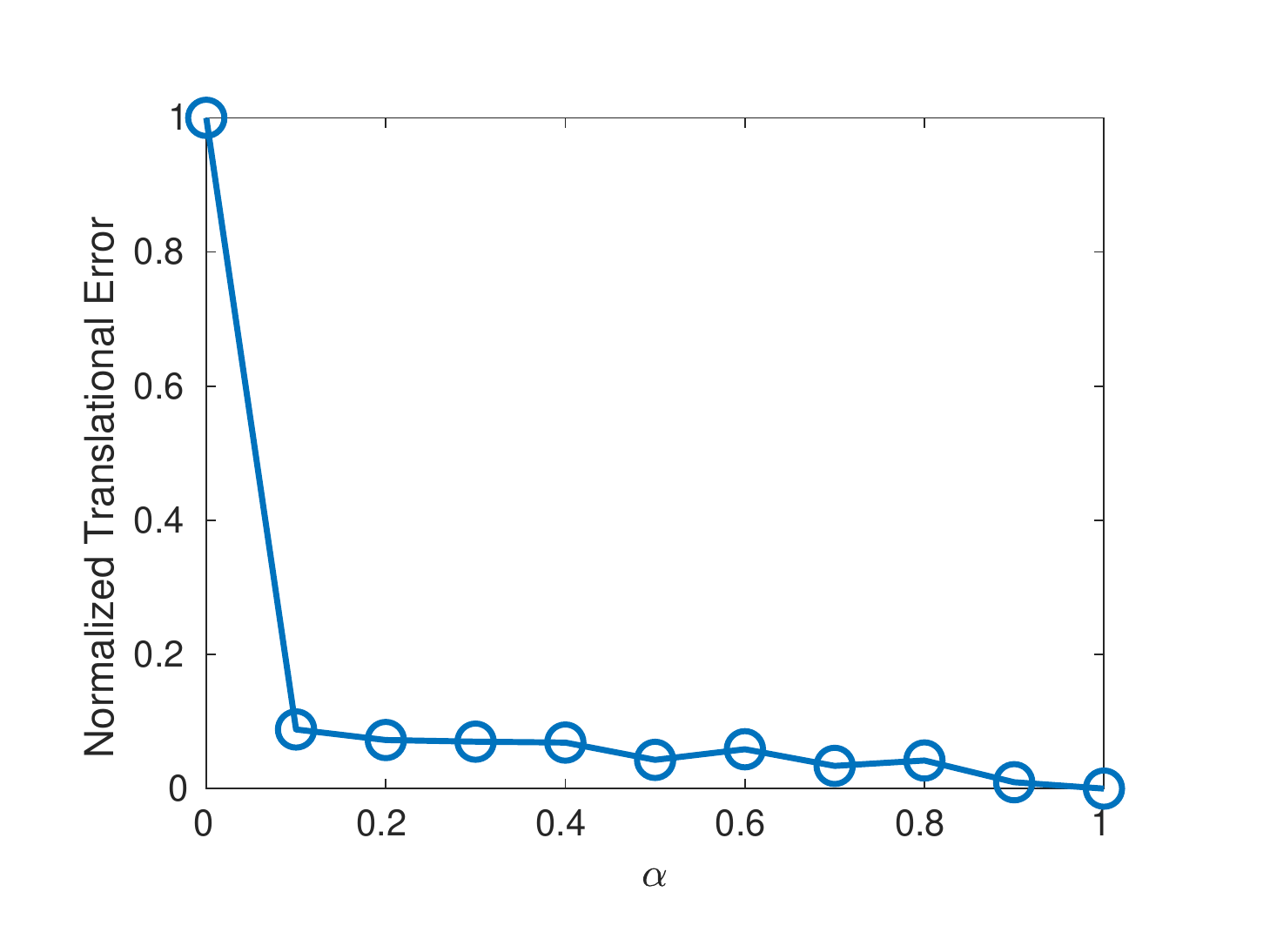}
    \includegraphics[width=4.1cm,trim=1cm .0cm 1cm .cm,clip]{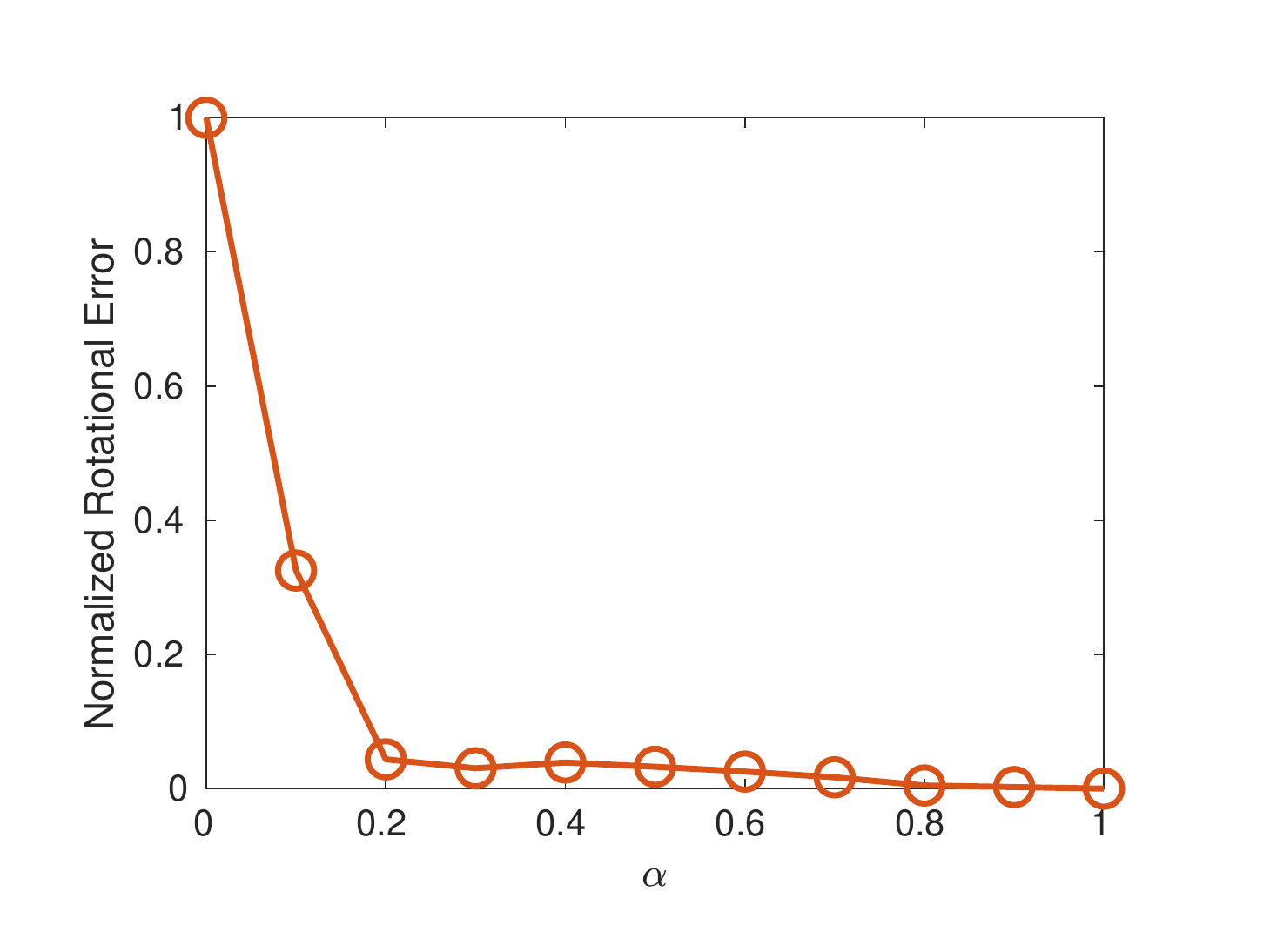}
    \vspace{-0.4cm}
    \caption{Normalized translation and rotation errors for different value of $ \alpha $.}
\label{fig:normalized_error_alpha}
\end{figure}

The bounded pose regression loss can blend together relative and composite transformation loss. However, it has been discovered by \cite{Clark2017} and confirmed in our experiments that training VO using composite transformation loss is difficult to converge due to the accumulative error of predictions. Fig. \ref{fig:normalized_error_alpha} shows normalized translation and rotation errors for different value of $\alpha$ in (\ref{eq:bounded_loss}) in the first training stage. It can be seen that training a DNN using only composite loss ($\alpha = 0$) leads to very large translation and rotation errors compared to when relative loss is also incorporated ($\alpha > 0$). The best performance is even achieved by training using relative loss only ($\alpha=1$), which indicates the difficulty in training with relative and composite losses right from the start. This motivates the utilization of Curriculum Learning (CL) where the learning process starts from the simplest objective and then increasing its difficulty. We refer to this mechanism as \textit{Geometry Aware Curriculum Learning} (GA-CL).

In the first stage of GA-CL, we start the training process by predicting a reasonable relative transformation (as suggested from Fig. \ref{fig:normalized_error_alpha}). This can be seen as minimizing the bounded pose regression loss from (\ref{eq:bounded_loss})-(\ref{eq:regular_loss}) with $\alpha = 1$. During the second stage, once the network has learned to produce reasonable relative transformations (as the validation loss no longer decreases), we may reveal more information to the network by gradually decreasing $\alpha$ so as to equalize relative and composite loss ($\alpha = 0.5$). In the final stage, we put more emphasize on the composite loss $0 < \alpha < 0.5$ such that the network can learn consistent composite transformation.

\begin{figure}[!ht]
    \centering
    \includegraphics[width=7.5cm,trim=0.1cm .9cm .1cm .cm,clip]{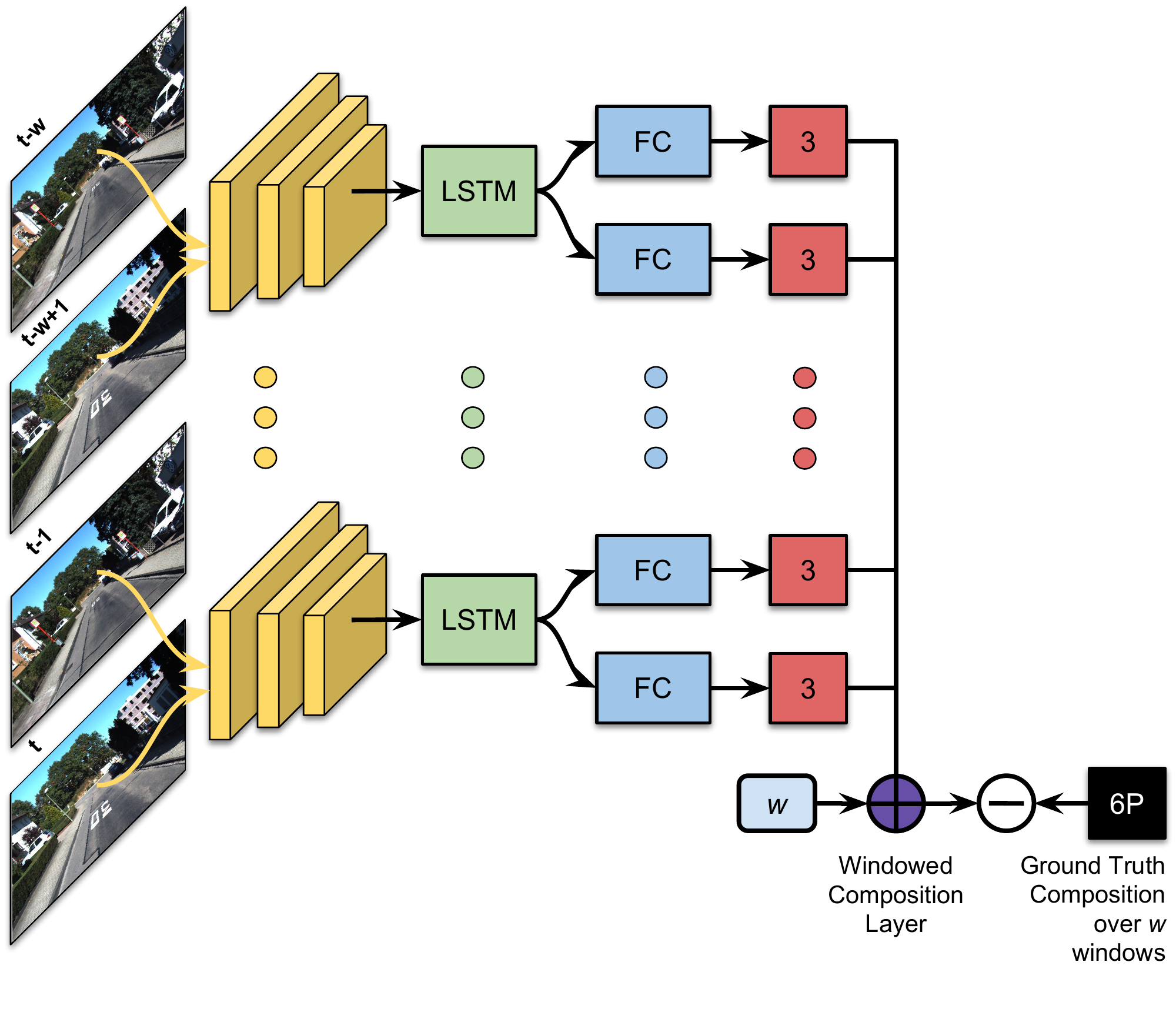}
    \caption{CL-VO architecture with a windowed composition layer to integrate relative estimates over small windows $w$.}
\label{fig:clvo_arch_composite}
\end{figure}

\subsection{Network Architecture}

The network architecture, dubbed CL-VO, is depicted in Fig. \ref{fig:clvo_arch_rel} and is mainly composed of a feature extractor and a pose regressor. The feature extractor is essentially a CNN aimed to learn optical-flow like features for VO estimation. We construct a cascade optical flow network which refines optical flow estimation subsequently from the previous sub-network for providing more detail flow estimation. We adopt FlowNet2-C \cite{ilg2017b} for the 1st network and Flownet2-S \cite{ilg2017b} for the 2nd and 3rd network. For producing the latent variables that can be directly consumed by the pose regressor, we remove the refinement part for the last optical flow network. 

The pose regressor part consists of two recurrent layers, in particular two Long Short Term Memory (LSTM) \cite{hochreiter1997} layers, followed by fully connected layers to estimate 6-DoF camera poses. Compared to directly using a fully connected layer for pose regressor, as seen in \cite{Muller2017} and \cite{Kendall2015}, LSTM is more suitable to learn the long term dependencies of camera pose since it can maintain its hidden state over time. The LSTM  operation can be formulated as follows:
\begin{align}
\label{eq:sparse_lstm_op}
\begin{bmatrix} \textbf{i} \\ \textbf{f} \\ \textbf{o} \\ \textbf{g} \end{bmatrix} & = \begin{bmatrix} \text{sigm} \\ \text{sigm} \\ \text{sigm} \\ \text{tanh} \end{bmatrix} \textbf{W}_{\text{lstm}}^{(l)} \begin{bmatrix} \textbf{h}_{t}^{(l-1)} \\ \textbf{h}_{t-1}^{(l)} \end{bmatrix}, \; \\
  \textbf{c}_{t}^{(l)} & = \textbf{f} \odot \textbf{c}_{t-1}^{(l)} + \textbf{i} \odot \textbf{g}, \; \\  
  \textbf{h}_{t}^{(l)} & = \textbf{o} \odot \text{tanh}(\textbf{c}_{t}^{(l)}), \;
\end{align}
where $\textbf{W}^{(l)}_{lstm} \in {\rm I\!R}^{4n^{(l)} \times (n^{(l-1)}+n^{(l)})}$ is the weight matrix for layer $l$, $n$ is tensor dimension, $t = 1,...,T$ is the timestep, and the vector $\textbf{h}_{t}^{(l)} \in {\rm I\!R}^{n^{(l)}}$ is its hidden state at step $t$ and layer $l$. Vector $\textbf{h}_{t}^{(0)}$ is equal to the input $\textbf{x}_{t}$ at step $t$. Operators \textit{sigm}, \textit{tanh}, and $\odot $ denote sigmoid function, hyperbolic tangent, and element-wise multiplication respectively.

For composing the relative transformation from a certain number of previous frames, we construct a differentiable custom \textit{windowed composition layer} as seen in Fig. \ref{fig:clvo_arch_composite}. A windowed composition layer concatenates the current frame-to-frame camera ego motion with the previous ego motion for a predefined number of window $w$ as follows
\begin{align}
\label{eq:windowed_composition}
\textbf{\^{p}}^{t}_{w} & = \textbf{\^{p}}^{t}_{t-1} \oplus ... \oplus \textbf{\^{p}}^{t-w+1}_{t-w} \oplus \textbf{\^{p}}^{t-w}. \;
\end{align}

\section{Experiments}

\subsection{Datasets}
Three datasets, consist of two public datasets and one self-collected dataset, are used in our experiments. The first dataset is KITTI autonomous driving dataset \cite{Geiger2012a}, a well-known public dataset for evaluating VO and SLAM algorithms. We use KITTI odometry data Sequences 00-10 for quantitative evaluation and Sequences 11-21 for qualitative evaluation. Although the dataset provides stereo imagery, we only use the left image for testing monocular VO algorithms. The second dataset is the Malaga urban dataset \cite{blanco2014}, which is also collected in a driving scenario. This dataset is only used to test a pre-trained model without training or fine-tuning. Similar to KITTI, we only utilize the left camera for testing monocular VO methods.

The last dataset is our self-collected human motion data imitating firefighter walking pattern. This dataset is collected in an indoor environment that consists of a corridor and a large room for approximately 1.5 hours. We use uEye global shutter camera mounted in a helmet, with VGA resolution  ($640 \times 480$) which runs at 30 Hz. The ground truth is taken from a ViCon Motion Capture system with approximately 1cm accuracy. The firefighter walking motion contains sweeping hand and foot for inspecting obstacles in front of the user, which is very challenging for monocular VO since it creates a zigzag motion pattern. Moreover, the moving hand occasionally obstructs some parts of the image.

\subsection{Competing Approaches}

To evaluate the performance of CL-VO, we compare our method with the state-of-the-art feature-based and learning-based VO methods, namely VISO2 \cite{Geiger2011}, ORB-SLAM \cite{Mur-Artal2015b}, and DeepVO \cite{Wang2017}. For VISO2, we use the monocular version (VISO2-M) for quantitative evaluation while we utilize the stereo version (VISO2-S) for qualitative comparison. We set the height of the camera in VISO2-M as described on each dataset paper to estimate the scale of the prediction. For ORB-SLAM, we used the result from \cite{Wang2017a} for quantitative evaluation. As for DeepVO, we constructed the DeepVO model with the same architecture and parameters as described in the paper. For each dataset, we trained DeepVO with the same settings as CL-VO (e.g. total training sequences, validation data, total epochs, optimizer, learning rate, etc.). We also train DeepVO with GA-CL to see how much improvement GA-CL can bring to DeepVO. 

\subsection{Implementation and Augmentation}

We implemented CL-VO using Tensorflow and Keras, and ran the training code on a NVIDIA TITAN V GPU. Before training, we computed the dataset mean and used it to normalize the image intensity. In order to provide more trajectory variations, we generated sequences with random start and end points, and random lengths. In every epoch, we constructed 10 random trajectories for each training sequence. The training can extend to 200 epochs for each training stage which takes around 10 hours, or can be stopped earlier if the validation loss shows no improvement.  We used the Adam optimizer with $1e-3$ as the initial learning rate. We also applied Dropout \cite{srivastava2014dropout} with $0.2$ dropout rate for regularizing the network. For parameter in (\ref{eq:bounded_loss})-(\ref{eq:regular_loss}), we set $[ \delta ; \zeta] =  [ 1; 100 ]$ for the KITTI dataset, and  $[ \delta ; \zeta] =  [1; 0.001]$ for the human motion dataset. For GA-CL setting, we mostly set the window $w=2$ or 3 and $ \alpha = 1$ for the 1st stage, $\alpha = 0.5$ for the 2nd stage, and $\alpha =  0.1$ for the 3rd stage as it get the best performance in KITTI dataset.

\begin{figure*}[!ht]
    \centering
    \subfloat[Estimated trajectory from Sequences 05 and 07]{
        \begin{tabular}[b]{c}%
        \includegraphics[width=4.2cm,trim=1cm .5cm 1cm .5cm,clip]{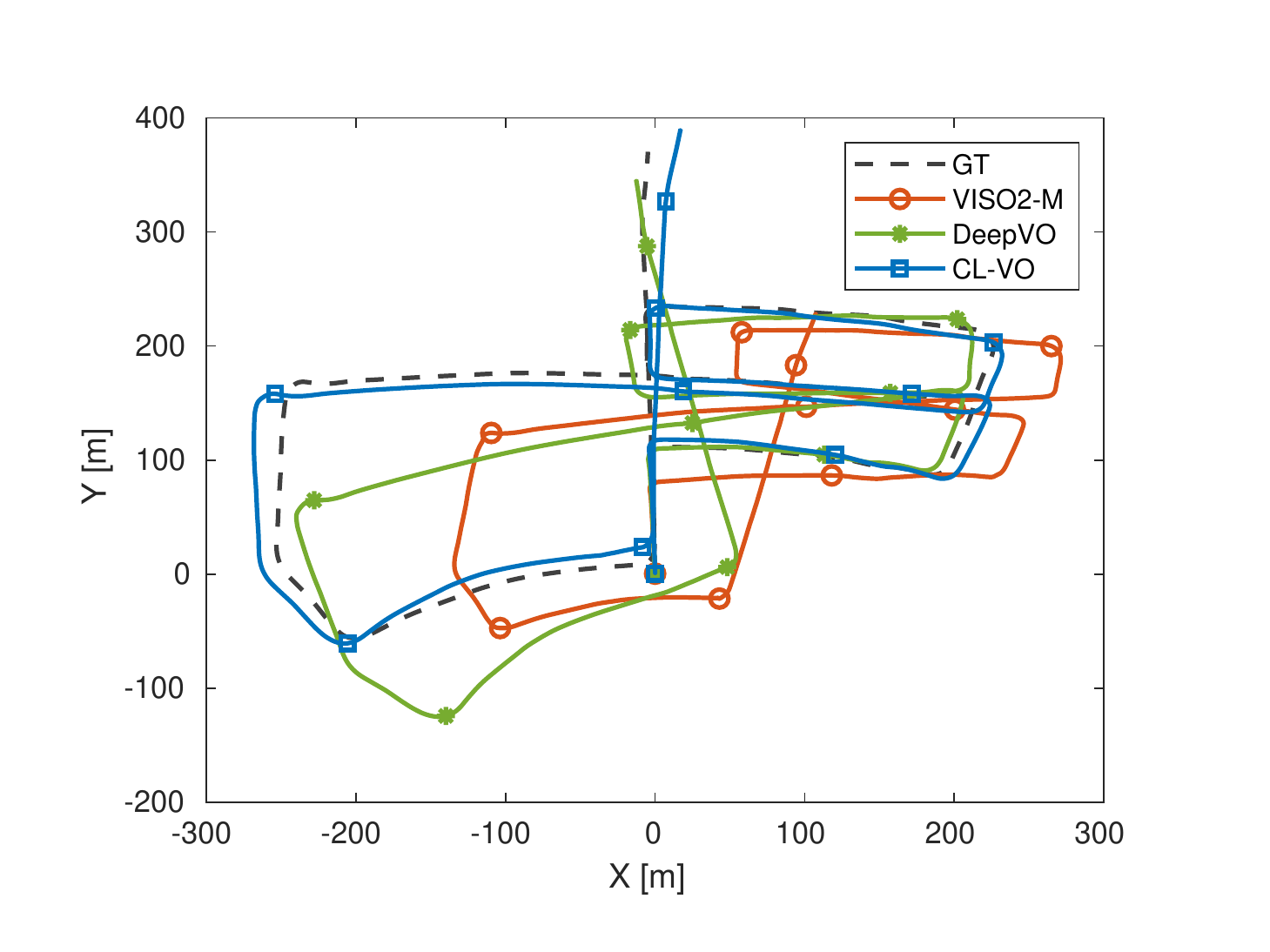}
        \hspace{-0.3cm}
        \includegraphics[width=4.2cm,trim=1cm .5cm 1cm .5cm,clip]{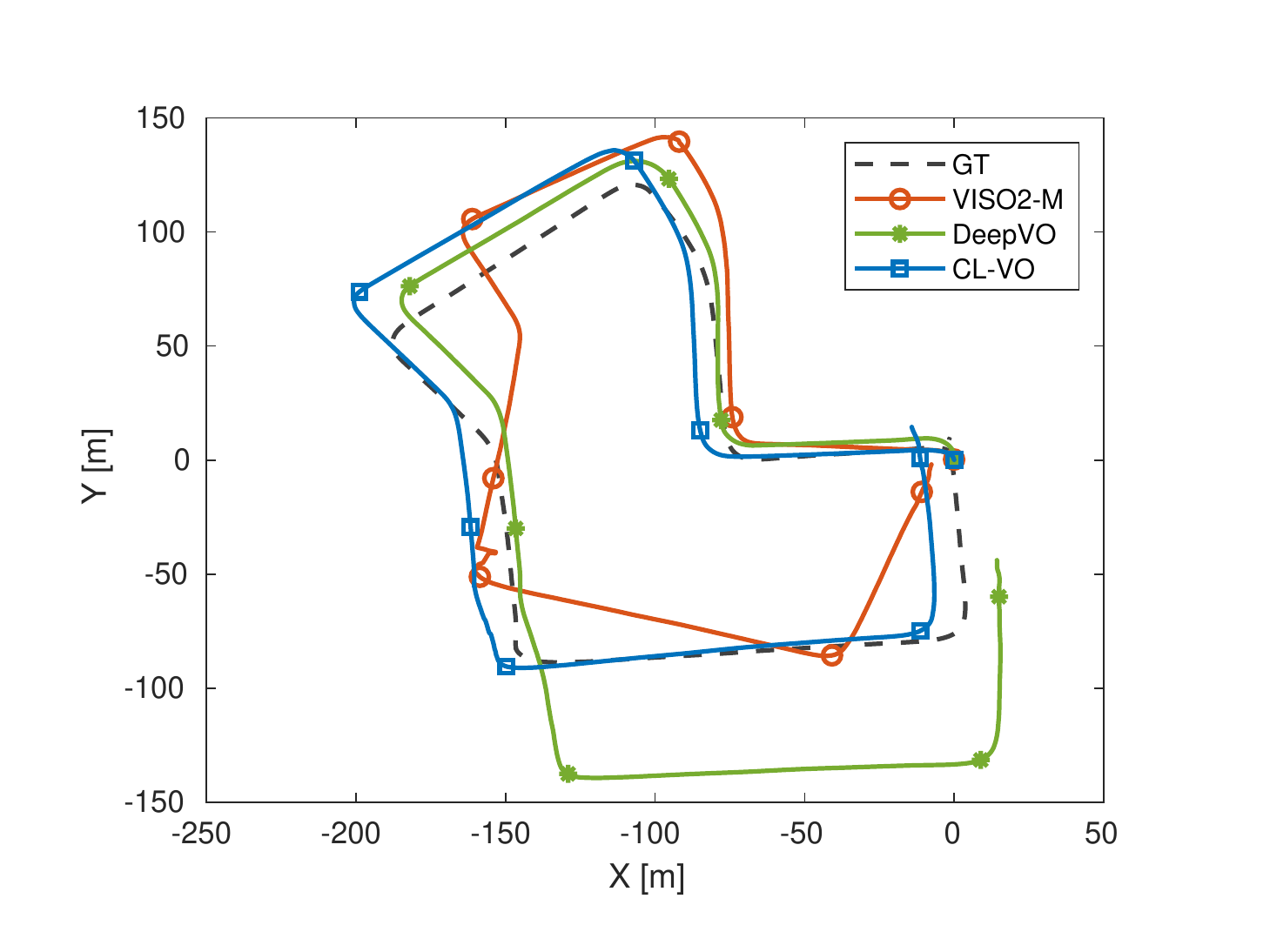}
        \end{tabular}
        }
    \subfloat[Estimated trajectory from Sequences 11 and 18]{
        \begin{tabular}[b]{c}%
    	\includegraphics[width=4.2cm,trim=1cm .5cm 1cm .5cm,clip]{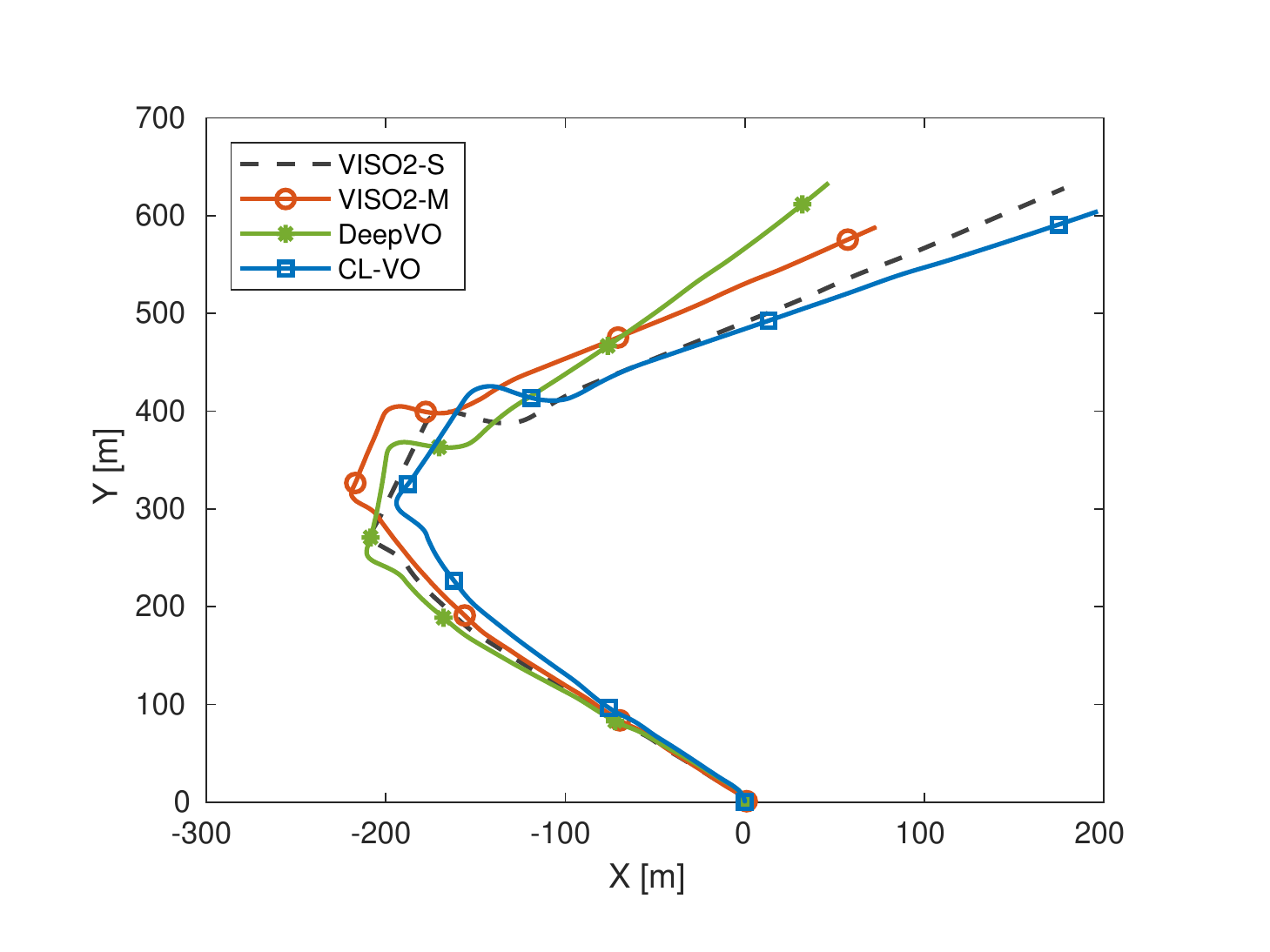}
    	\hspace{-0.3cm}
        \includegraphics[width=4.2cm,trim=1cm .5cm 1cm .5cm,clip]{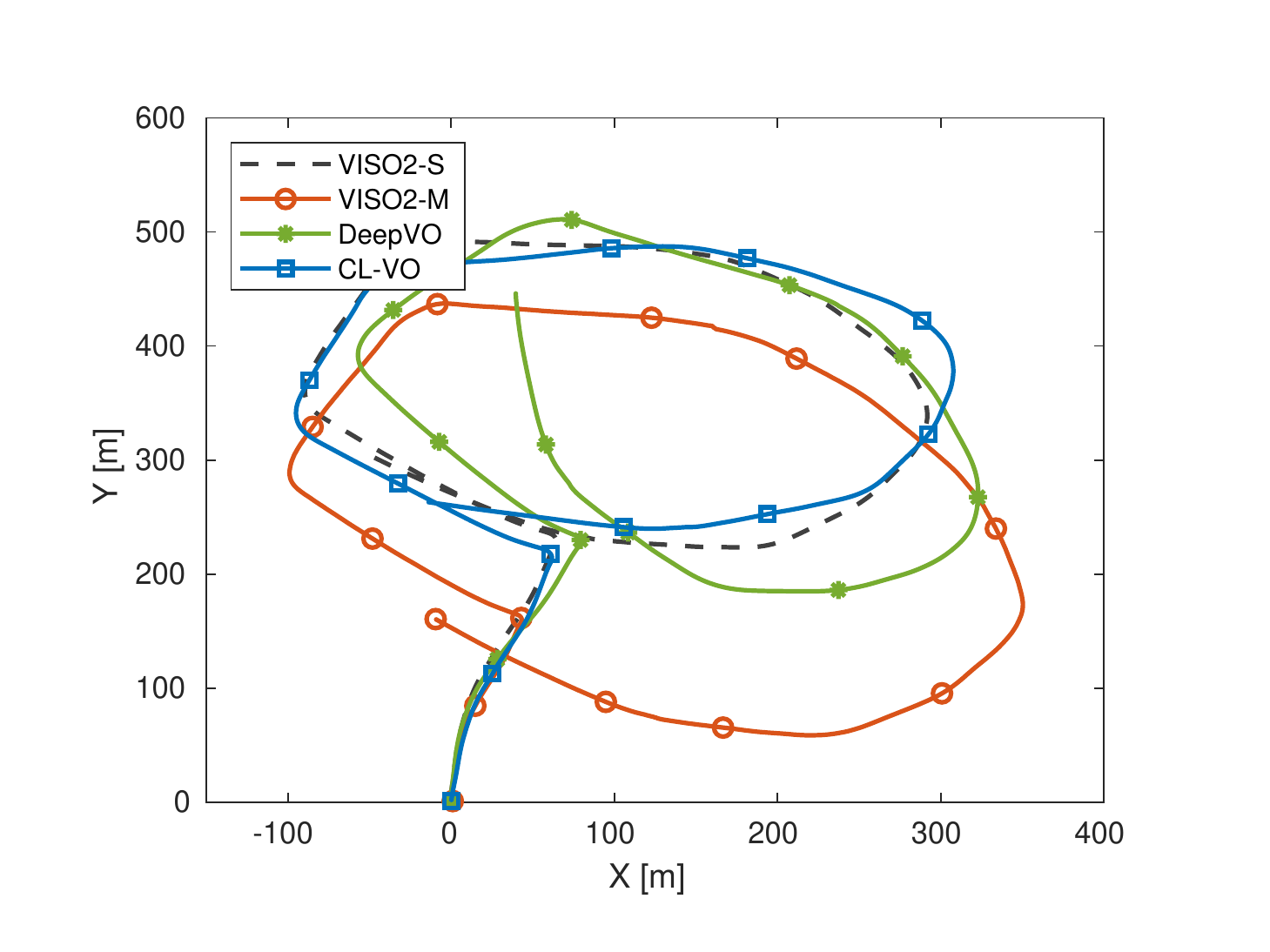}
        \end{tabular}
        }
    \caption{ (a) Qualitative results from Sequences 05 and 07 on KITTI dataset. (b) Qualitative results from Sequences 11 and 18 on KITTI dataset. Note that the ground truth is not available for KITTI Sequences 11-20.}
\label{fig:traj_kitti}
\end{figure*}

\subsection{Results}

\subsubsection{Tests on KITTI Dataset}

We performed two experiments on the KITTI dataset. The first experiment is conducted for KITTI Sequences 00-10 where precise ground truth is available such that quantitative evaluation can be conducted. The second experiment is aimed to test further the generalization of the network on KITTI testing Sequences 11-20. Since there is no ground truth available for KITTI Sequences 11-20, no quantitative evaluation is performed. 

\begin{figure}[!ht]
    \centering
    \includegraphics[width=4.1cm,trim=0.5cm .5cm 1cm .5cm,clip]{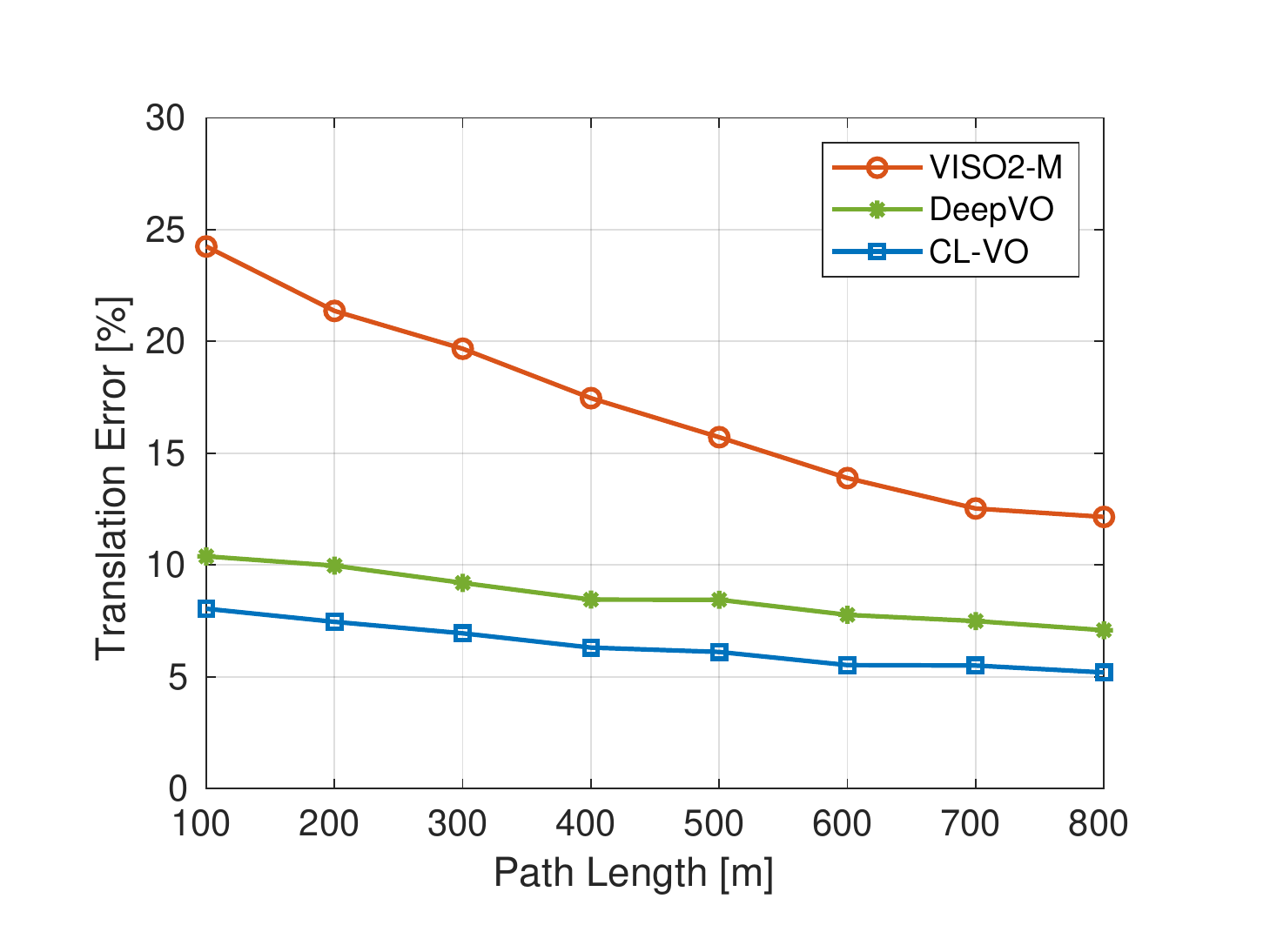}
    \includegraphics[width=4.1cm,trim=0.5cm .5cm 1cm .5cm,clip]{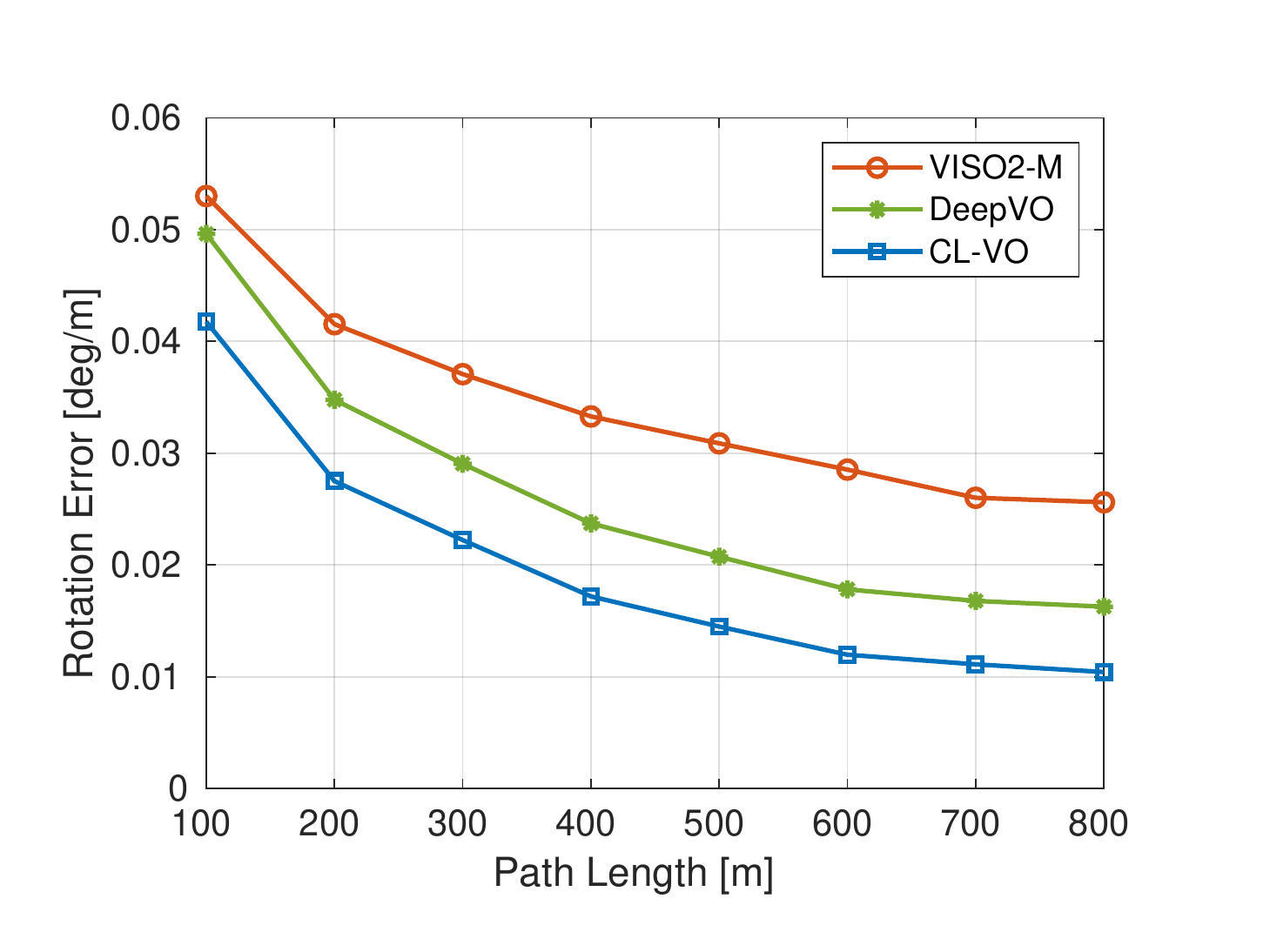} \\
    \vspace{-0.2cm}
    \caption{Translation and rotation errors against path length on KITTI dataset.}
\label{fig:trans_rot_vs_length}
\end{figure}

For the first experiment, we trained CL-VO on KITTI Sequences 00, 01, 02, 08, and 09, and tested on KITTI Sequences 03, 04, 05, 06, 07, and 10 as seen in \cite{Wang2017}. Fig. \ref{fig:traj_kitti} (a) shows the qualitative results from Sequences 05 and 07. It can be seen that all CL-VO predictions are relatively accurate and consistent against the ground truth. CL-VO significantly outerforms VISO2-M and DeepVO. As for VISO2-M, the VO estimation in Fig. \ref{fig:traj_kitti} (a) suggest that the scale estimation using fixed camera height is not robust against noise due to car jolts during driving \cite{Wang2017a}. Note that neither scale estimation nor post alignment to ground truth is conducted for CL-VO. The quantitative results can be seen in Fig. \ref{fig:trans_rot_vs_length} where CL-VO consistently yields better performance for both translation and rotation against the path length compared to VISO2-M and DeepVO. Table \ref{table:rpe_errors} details the frame-to-frame relative transformation errors of the compared algorithms for each testing sequences. The result indicates that CL-VO achieves more robust outputs than VISO2-M, ORB-SLAM, and DeepVO, although the performance is, as expected, worse than the stereo algorithm, i.e. VISO2-S. The table also shows that GA-CL can boost the performance of DeepVO by $21 \%$ and $16 \%$ for translation and rotation respectively. CL-VO achieves higher accuracy than DeepVO+GA-CL as it estimates more accurate optical flow through the cascade optical flow networks.

For the second experiment, we trained CL-VO on KITTI Sequences 00-10 and tested on KITTI testing Sequences 11-20. Qualitatively, we can see from Fig. \ref{fig:traj_kitti} (b) that CL-VO predictions are more similar to the stereo algorithm (VISO2-S) estimation than VISO2-M and DeepVO. This confirms that CL-VO can generalize well in new scenarios with different motion patterns and environments although it suffers from drift over time.

\setlength{\tabcolsep}{4pt}
\begin{table*}
\begin{center}
\caption{Frame-to-frame relative translation and rotation errors on KITTI dataset. }
\label{table:rpe_errors}
\begin{tabular}{c|cccccccccc|cc}
\noalign{\smallskip}
\noalign{\smallskip}
\noalign{\smallskip}
& \multicolumn{10}{c}{Monocular VO} & \multicolumn{2}{c}{Stereo VO} \\
\hline
Seq & \multicolumn{2}{c}{VISO2-M} & \multicolumn{2}{c}{ORB-SLAM} & \multicolumn{2}{c}{DeepVO} & \multicolumn{2}{c}{DeepVO+GA-CL  (\textbf{ours})} & \multicolumn{2}{c}{CL-VO (\textbf{ours})} & \multicolumn{2}{c}{VISO2-S} \\
\hline
& trans(\%) & rot($^{\circ}$) & trans(\%) & rot($^{\circ}$) & trans(\%) & rot($^{\circ}$) & trans(\%) & rot($^{\circ}$) & trans(\%) & rot($^{\circ}$) & trans(\%) & rot($^{\circ}$) \\
\hline
03 & 28.14 & \textbf{0.0230} & 21.07 & 0.1836 & 10.71 & 0.0479 & 8.36 & 0.0353 & \textbf{8.12} & 0.0347 & 3.21 & 0.0325 \\
04 & 33.92 & \textbf{0.0177} & \textbf{4.46} & 0.0560 & 9.95 & 0.0407 & 8.66 & 0.0308 & 7.57 & 0.0261 & 2.12 & 0.0212 \\
05 & 14.65 & 0.0397 & 26.01 & 0.3427 & 8.02 & 0.0265 & 5.81 & 0.0210 & \textbf{5.77} & \textbf{0.0200} & 1.53 & 0.0160 \\
06 & 19.54 & 0.0249 & 17.47 & 0.1717 & \textbf{7.10} & 0.0186 & 7.39 & 0.0183 & 7.66 & \textbf{0.0166} & 1.48 & 0.0158 \\
07 & 12.69 & 0.0647 & 24.53 & 0.3890 & 16.20 & 0.0380 & 9.79 & 0.0413 & \textbf{6.79} & \textbf{0.0300} & 1.85 & 0.0191 \\
10 & 30.39 & 0.0306 & 86.51 & 0.9890 & 9.04 & 0.0391 & 8.30 & 0.0303 & \textbf{8.29} & \textbf{0.0294} & 1.17 & 0.0130 \\
\hline
avg & 23.22 & 0.0334 & 30.01 & 0.3553 & 10.17 & 0.0351 & 8.05 & 0.0294 & \textbf{7.37} & \textbf{0.0267} & 1.89 & 0.0196 \\
\hline
\end{tabular}
\end{center}
\end{table*}
\setlength{\tabcolsep}{1.4pt}

\subsubsection{Generalization in Malaga Dataset}

In order to further test the generalization ability of the proposed framework, we  tested CL-VO on the Malaga dataset without any furter training or fine-tuning. We used the CL-VO model which is trained on KITTI dataset Sequences 00-10 and tested directly on the Malaga image data. Since the image resolution in the Malaga dataset is different from KITTI, we cropped the images to the KITTI image size. Some image information is expected to lost during this cropping process which might affect the final predictions.

\begin{figure*}[!ht]
    \centering
    \includegraphics[width=0.58\columnwidth]{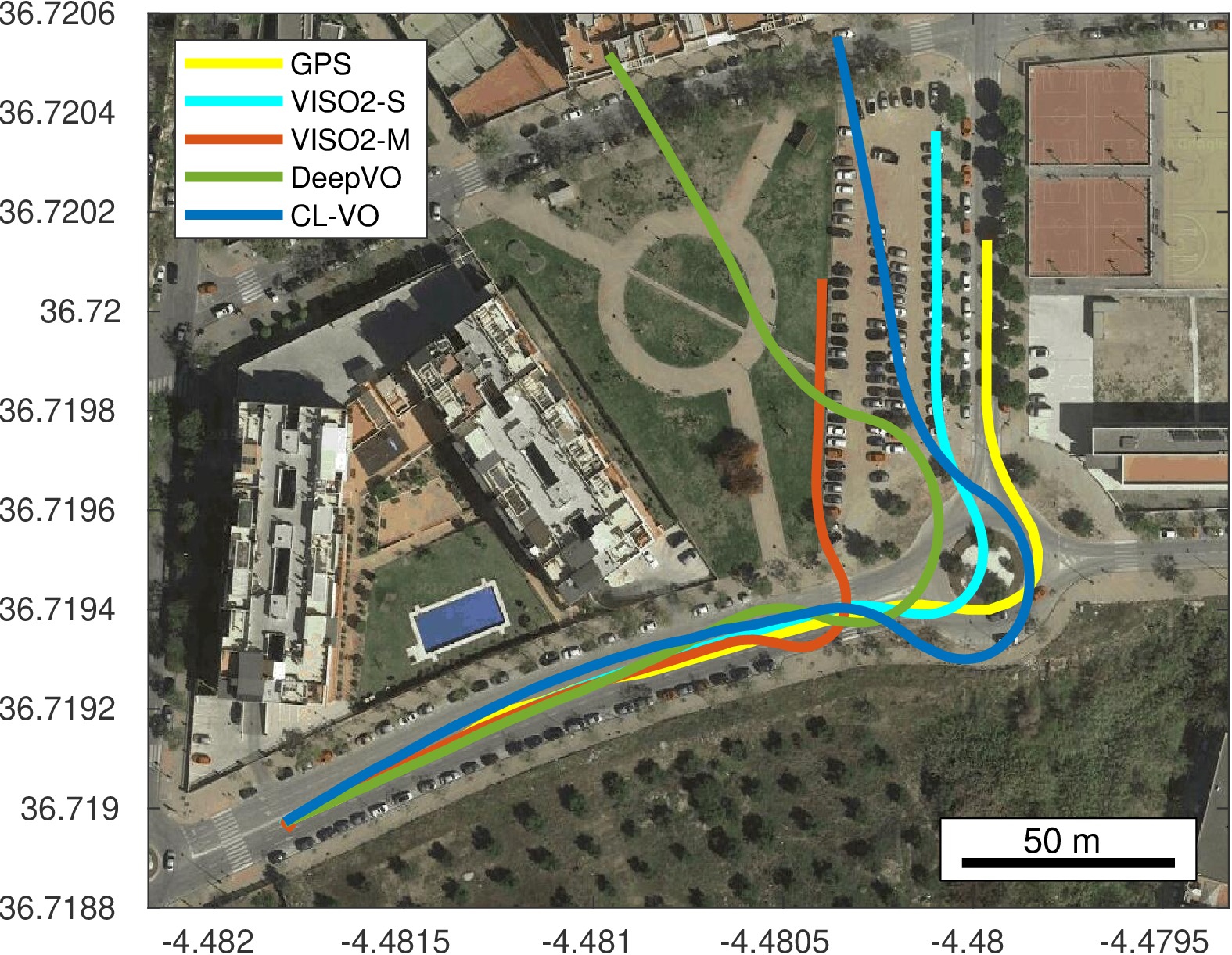}
    \includegraphics[width=0.58\columnwidth]{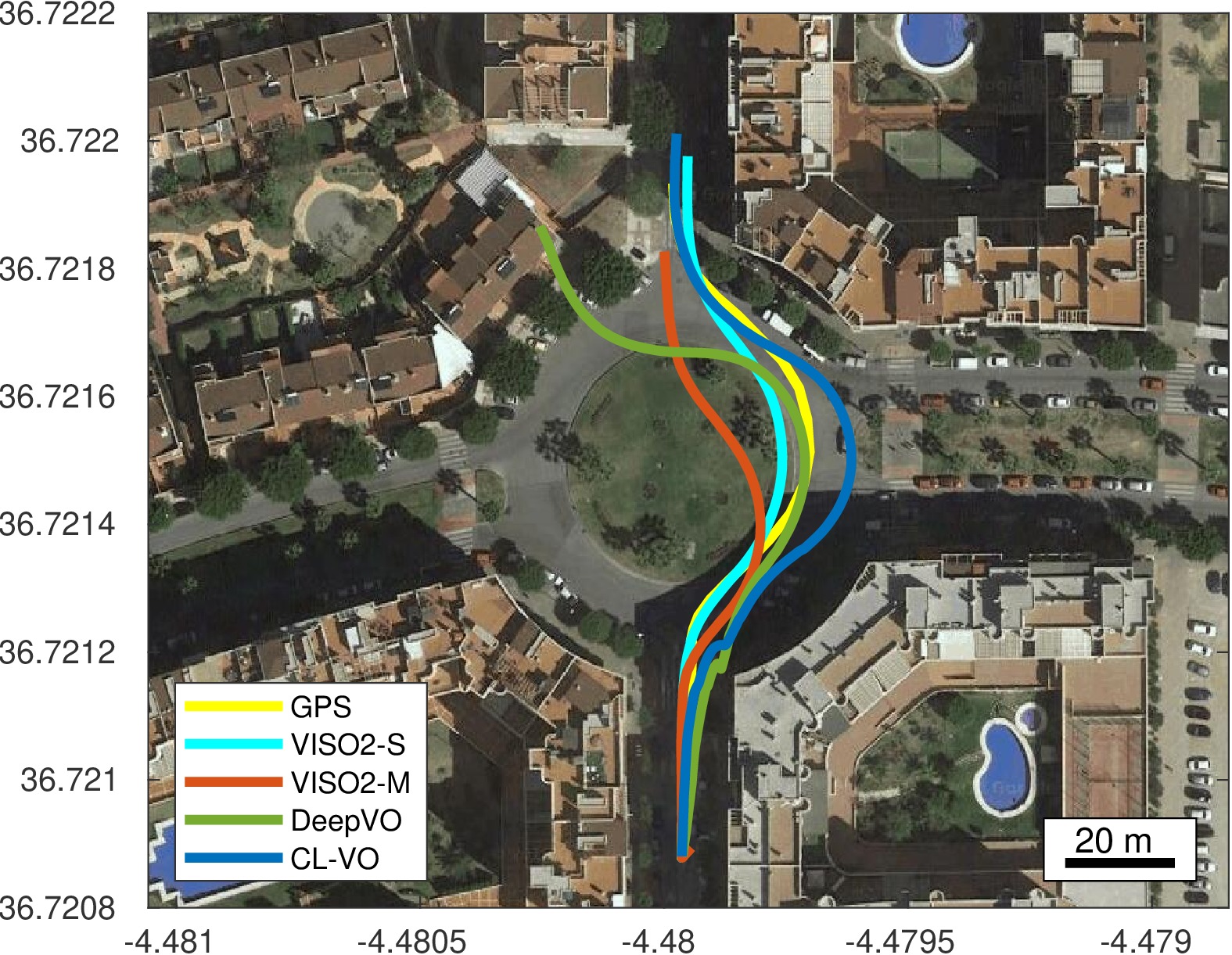}
    \includegraphics[width=0.605\columnwidth]{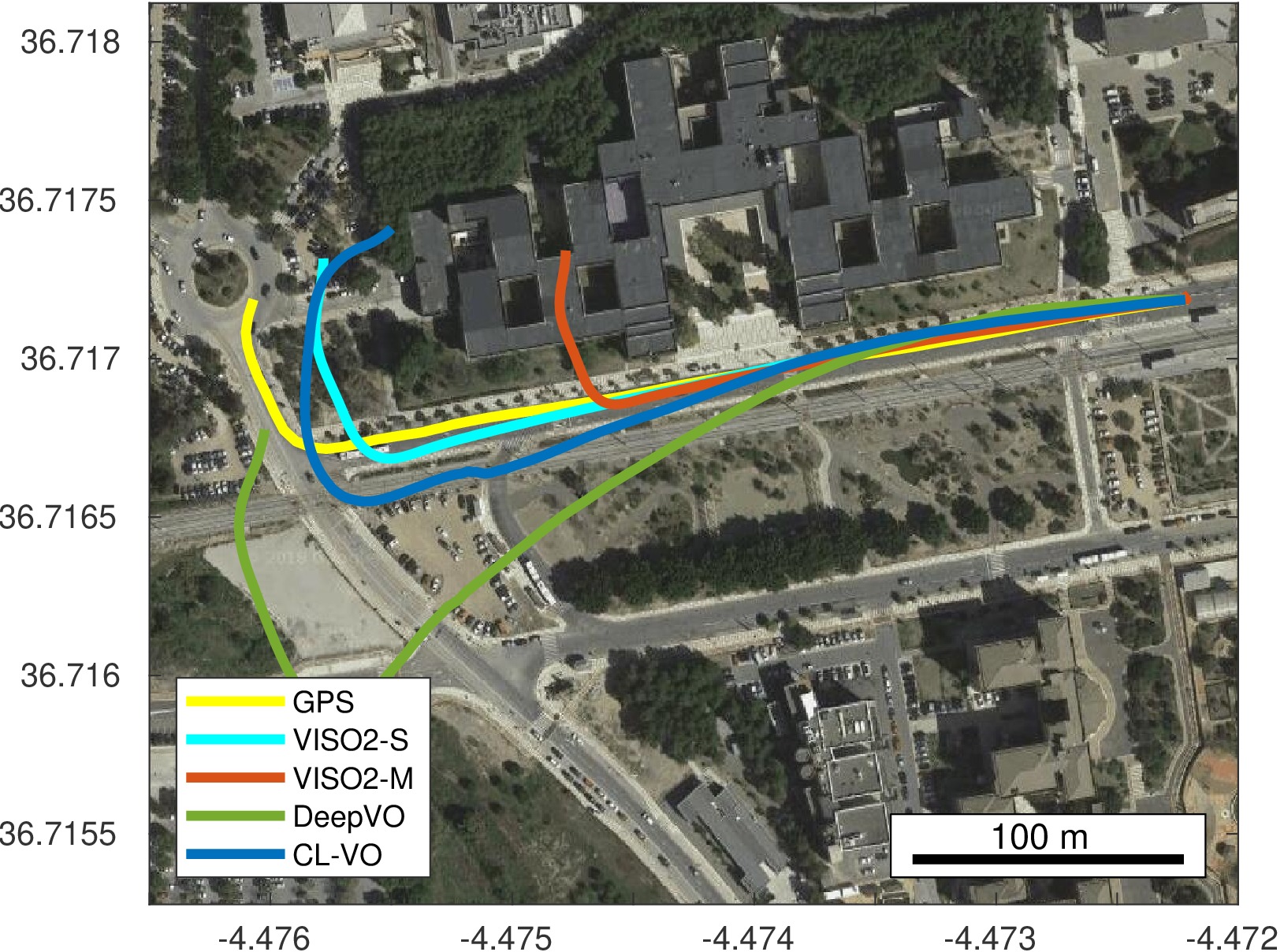}
    \caption{Generalization tests on Malaga Dataset superimposed on Google Map. DeepVO and CL-VO are only trained on KITTI dataset Sequences 00-10.}
\label{fig:test_malaga}
\end{figure*}

Fig. \ref{fig:test_malaga} depicts the test results on Malaga dataset Sequences 03, 04, and 09, superimposed on Google Map. Since the Malaga dataset does not have ground truth, a quantitative evaluation cannot be conducted. However, since frequent GPS data is available, we still can perform qualitative comparison. As we can see from Fig. \ref{fig:test_malaga}, CL-VO predictions are close to GPS and VISO2-S in those three sequences. It is significantly better than VISO2-M and DeepVO, although it suffers from drift. This experiment further confirms that CL-VO generalizes to other datasets which are collected with different cameras in different environments. This also shows that CL-VO generalizes better than DeepVO as the drift of DeepVO is larger on the test sequences. 

\subsubsection{Tests on Human Motion Dataset} We divided the human motion dataset into 2 groups, 1 hour and 15 minutes for training and the remaining 15 minutes for testing. We subsample one frame for every six images to provide enough displacement between consecutive frames.

\begin{figure*}[!ht]
    \centering
    \subfloat[Test on human walking data]{
        \begin{tabular}[b]{c}%
        \includegraphics[width=5.5cm,trim=1cm 0.7cm 1.3cm 1cm,clip]{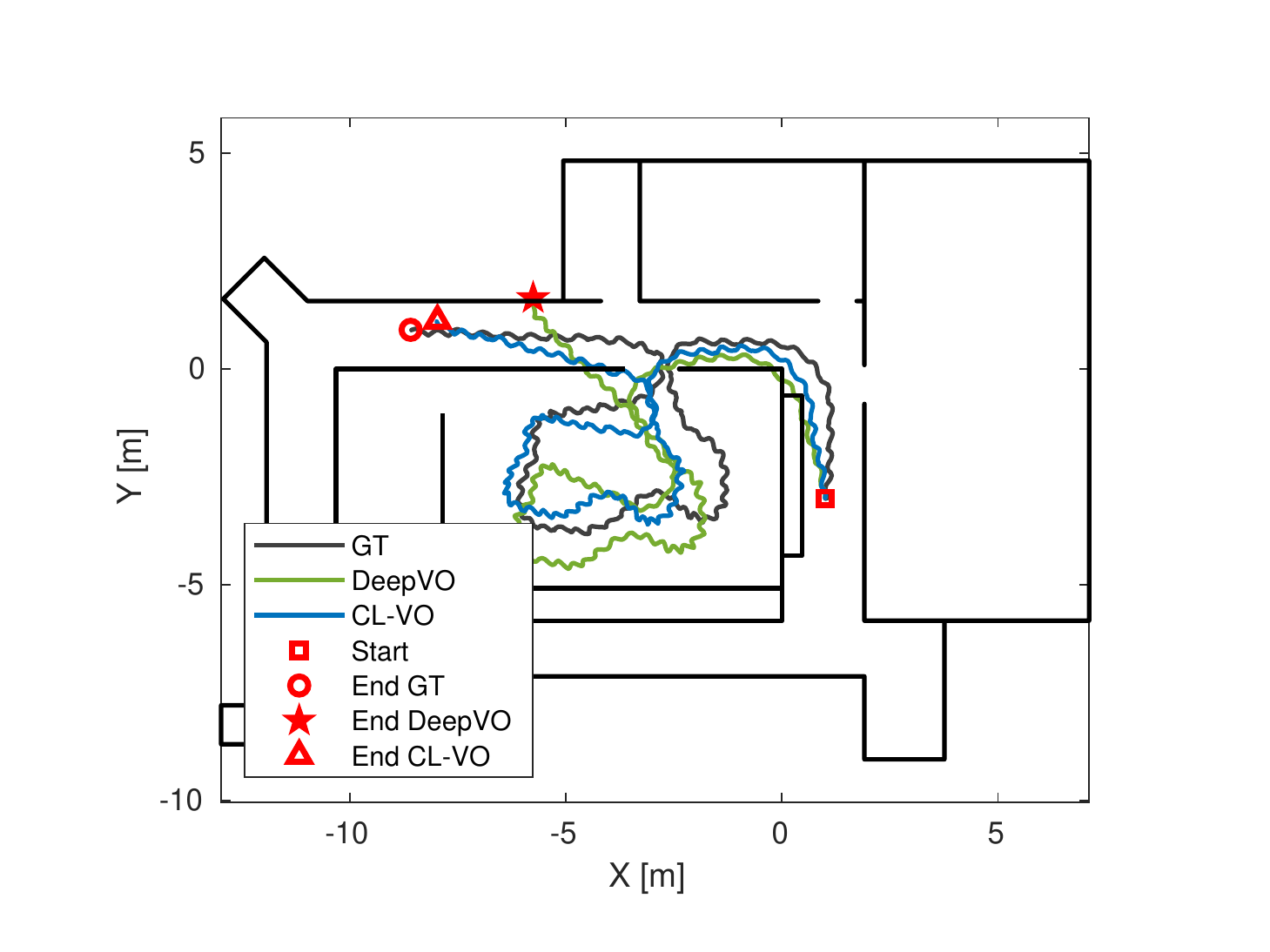}
        \end{tabular}
        }
        \hspace{-0.4cm}
    \subfloat[Estimates of 6-DoF camera poses]{
        \begin{tabular}[b]{c}%
    	\includegraphics[width=5.6cm,trim=1.6cm 1.5cm 1.5cm 1cm,clip]{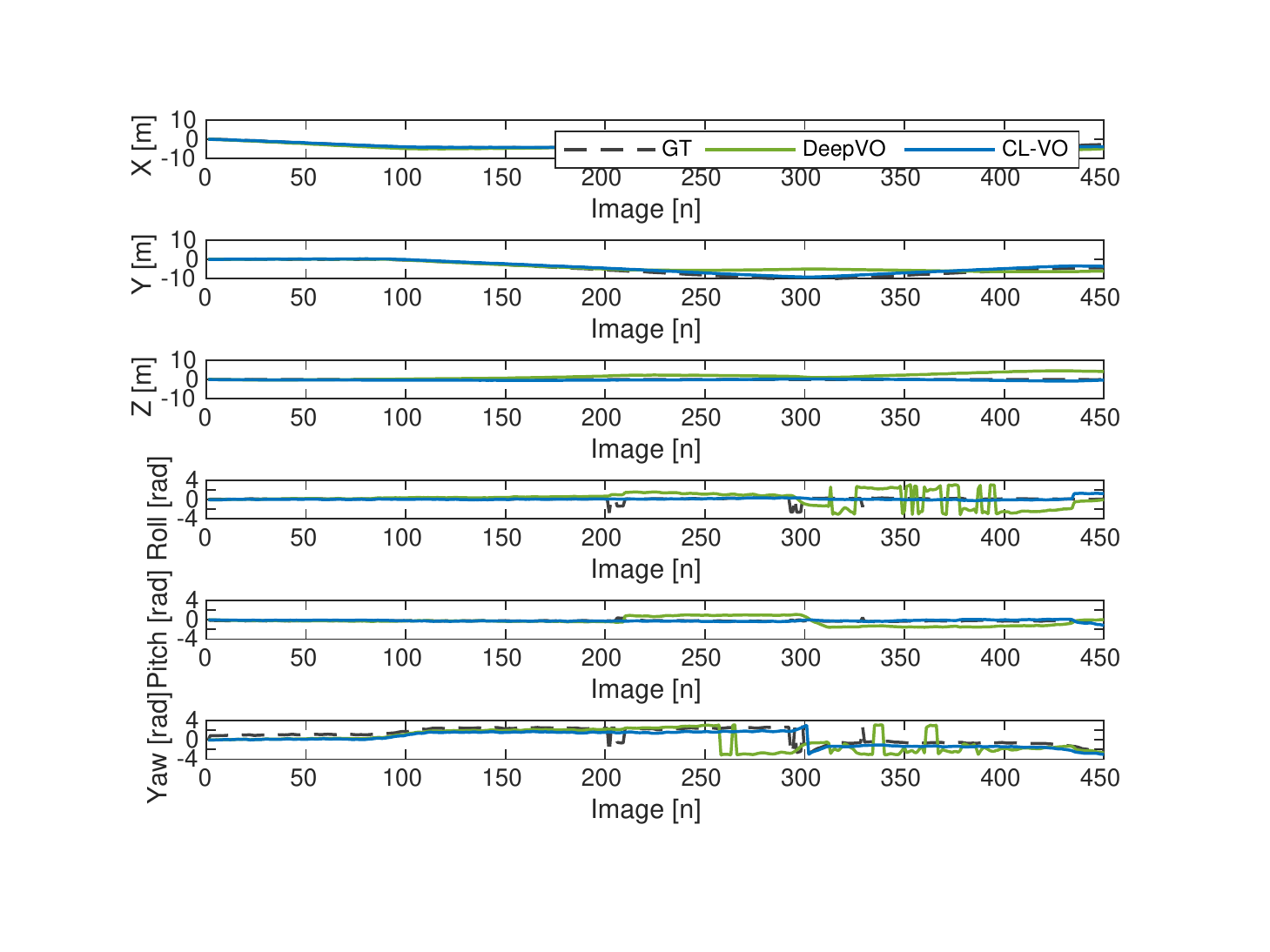}
        \end{tabular}
        }
        \hspace{-0.4cm}
    \subfloat[CDF errors]{
        \begin{tabular}[b]{c}%
    	\includegraphics[width=5.5cm,trim=1cm 0.7cm 1cm 1cm,clip]{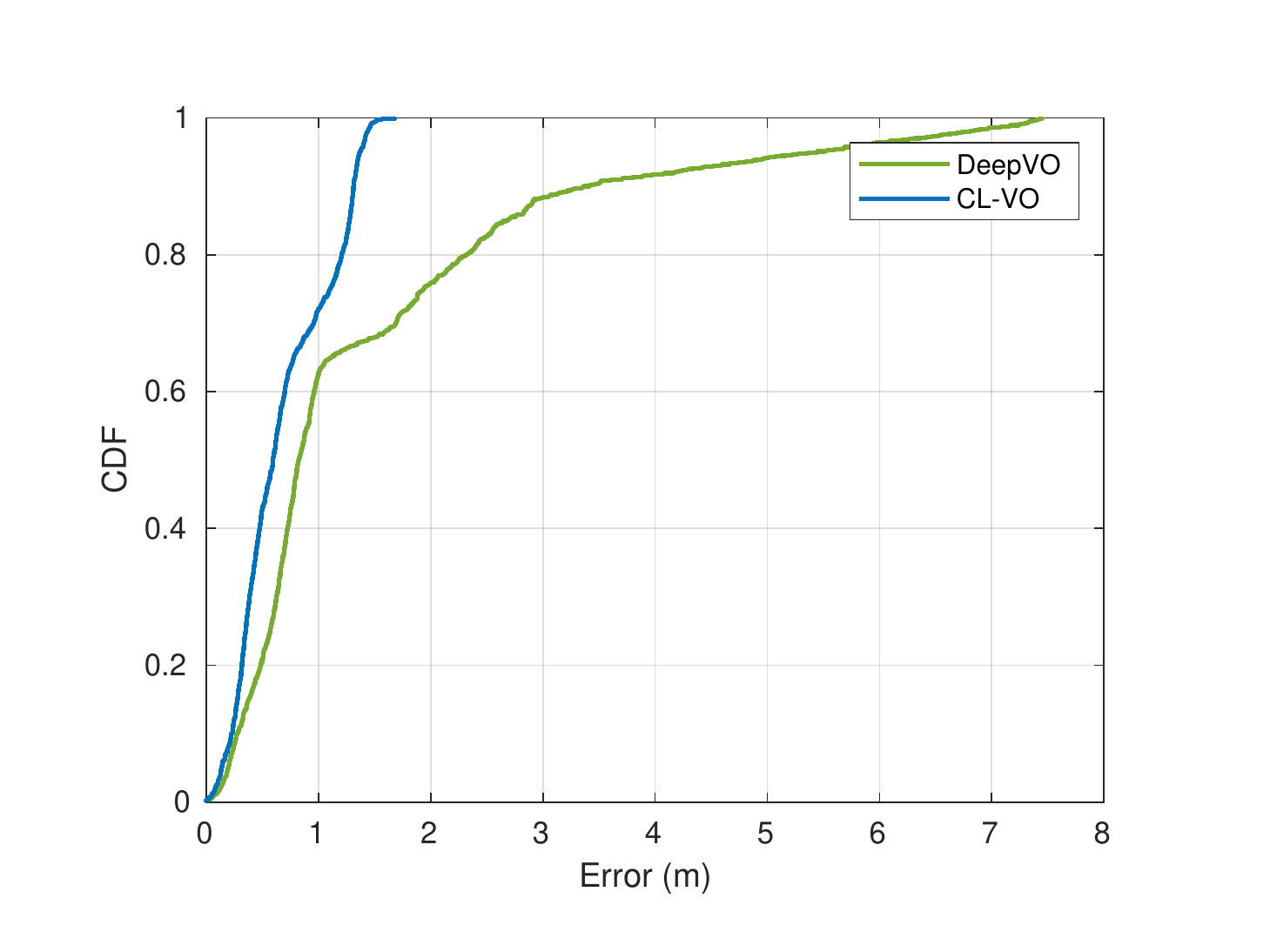}
        \end{tabular}
        }
    \caption{ (a) Test on human walking data in an office building. (b) 6-DoF camera poses compared to ground truth. (c) CDF of RMS absolute errors.}
\label{fig:traj_vicon_ff}
\end{figure*}

Fig. \ref{fig:traj_vicon_ff} (a) shows the qualitative results on one of the test sequences. It can be seen that CL-VO performs better than DeepVO as the prediction is closer to the ground truth. While CL-VO successfully tracks the camera movement, DeepVO fails to perform turning accurately which leads to much larger drift. Fig. \ref{fig:traj_vicon_ff} (b) shows the 6-DoF translation (x, y, z) and orientation (roll, pitch, yaw) of CL-VO compared with DeepVO and ground truth. It is clear that CL-VO tracks the changes on translation and orientation accurately. Fig. \ref{fig:traj_vicon_ff} (c) illustrates the distribution of the absolute errors (RMSE). CL-VO significantly outperform DeepVO, achieving less than 2 meters errors during $100 \%$ of testing time.

\subsubsection{The Impact of Geometry-Aware Curriculum Learning}

We performed an ablation study to understand the impact of the geometry-aware curriculum learning (GA-CL). We compare the performance of the proposed network when it is trained with the curriculum, reversed curriculum (anti-curriculum), and without curriculum. For training without curriculum, we use two loss functions, namely the standard relative loss and the bounded pose regression loss with $w=2$ and $\alpha = 0.5$. For the anti-curriculum, the stages described in Section \ref{sec:gacl} are reversed. All competing networks are trained with the same setting except GA-CL and anti-curriculum changes the parameter of the objective function at the end of each training stage.

\begin{figure}[!ht]
    \centering
    \includegraphics[width=4.25cm,trim=1cm 1cm 1.2cm 0.5cm,clip]{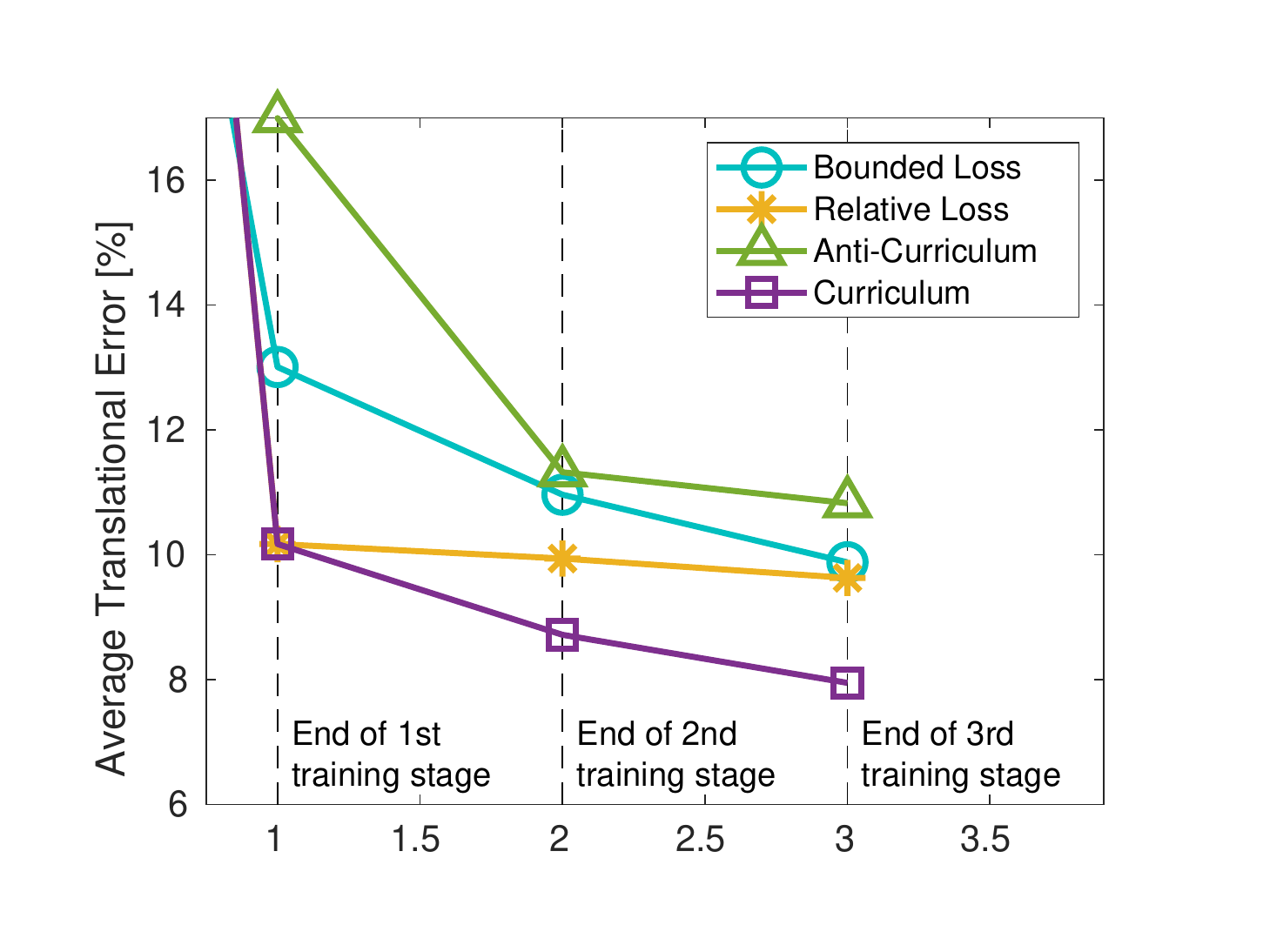}
    \hspace{-0.3cm}
    \includegraphics[width=4.1cm,trim=1cm 1cm 1cm 0.5cm,clip]{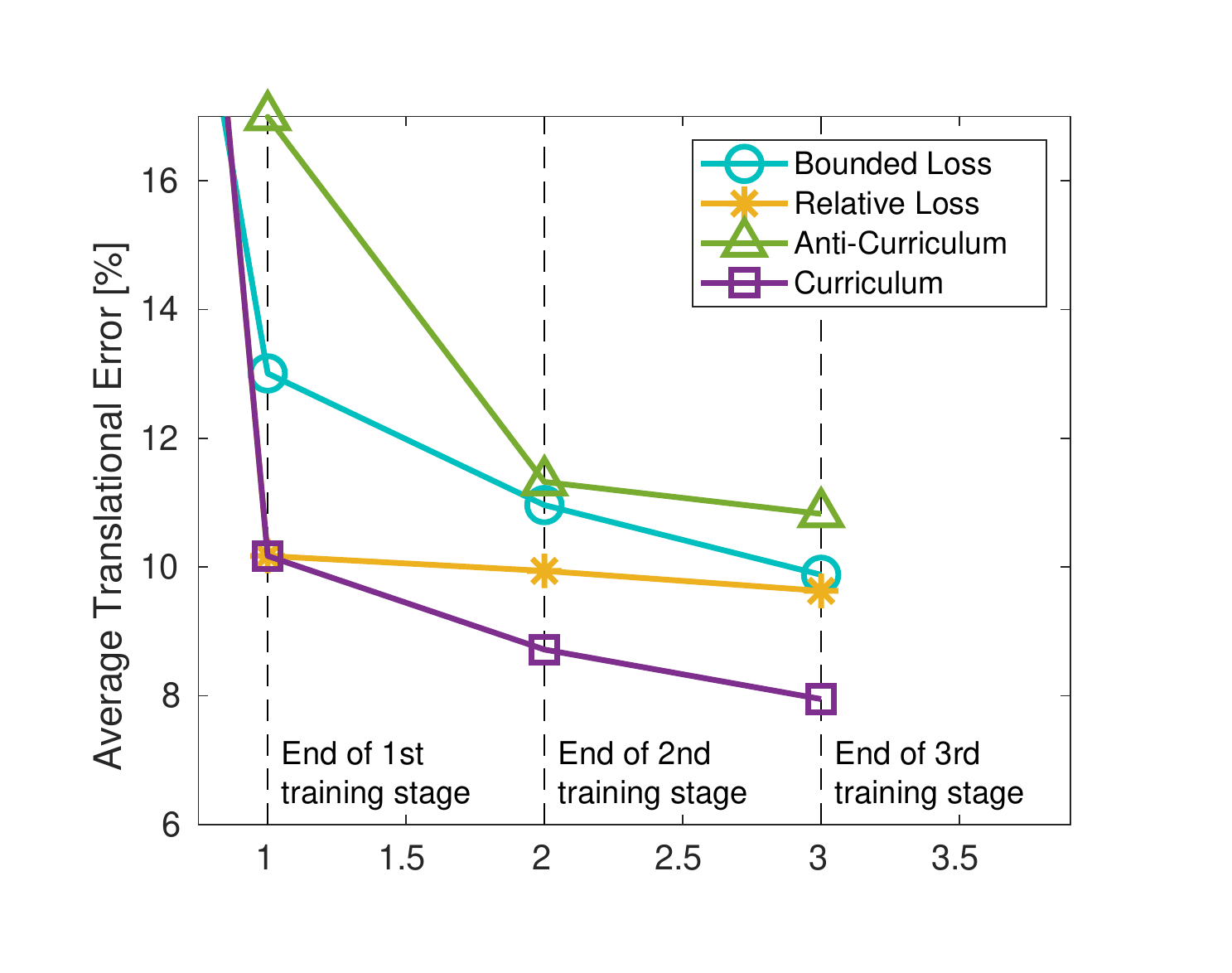}
    \caption{The impact of GA-CL algorithm on translation and rotation errors.}
\label{fig:ablation}
\end{figure}

Fig. \ref{fig:ablation} depicts the key results of this study. As expected, directly training the network with the bounded loss is more difficult to converge although the performance gradually improves in later stages of training. On the other hand, the network trained with the relative loss already reaches a stable state in the first stages of training. It only improves slightly afterwards or can even lead to overfitting as the accuracy of the rotation part decreases. The anti-curriculum gets very low accuracy in the beginning although the performance is improving after training with relative loss. Finally, the network trained with GA-CL can converge and generalize better which results in significantly lower translation and rotation errors in each training stages.

One possible explanation for this performance gain is GA-CL can be regarded as a special form of transfer learning, where the initial tasks (minimizing relative transformation loss) are used to guide the learner such that it can perform better on the final task (minimizing bounded pose regression loss). While the motivation of conventional transfer learning is to improve the generalization by sharing model weights across tasks, GA-CL introduces the idea of guiding the optimization process, either for faster convergence or better local minima \cite{bengio2009}. Another perspective is GA-CL can be seen as a way to gradually injecting domain knowledge into DNNs by progressively reveals more information to the network over time via objective function alteration.

\section{Conclusion}
In this paper, we have presented a novel DNN framework (CL-VO) which is trained using a geometry-aware objective function and curriculum learning (GA-CL). We have shown that CL-VO performed significantly better than state-of-the-art feature-based and learning-based approaches. We have also shown that GA-CL strategy can significantly improve the generalization ability of the network for both translation and rotation components, compared to a network that is trained without GA-CL. We believe that CL-VO can be a viable complement to conventional VO approaches.

\textit{Acknowledgement}. This research is funded by the US National Institute of Standards and Technology (NIST) Grant No. 70NANB17H185. Muhamad Risqi U. Saputra was supported by Indonesia Endowment Fund for Education (LPDP).

\addtolength{\textheight}{-8cm}   






\bibliographystyle{IEEEtran}

\end{document}